\documentclass{article}

\usepackage{microtype}
\usepackage{graphicx}
\usepackage{subfigure}
\usepackage{booktabs}
\usepackage{amsmath,amssymb}
\usepackage{xcolor}
\usepackage{multirow}
\usepackage{tikz}
\usetikzlibrary{arrows.meta,positioning,calc,shapes.geometric}
\usepackage{hyperref}

\usepackage[accepted]{mlsys2025} 
\makeatletter\renewcommand{\Notice@String}{Preprint.}\makeatother 
\usepackage[T1]{fontenc}   
\usepackage{algorithm}
\usepackage{algorithmic}

\newcommand{\todo}[1]{\textcolor{red}{[TODO: #1]}}

\definecolor{gaingreen}{HTML}{1A7F37}\definecolor{lossred}{HTML}{B42318}   

\usepackage{comment}\usepackage[normalem]{ulem}
\newif\ifshowedits \showeditstrue
\definecolor{editblue}{HTML}{1F4FBF}\definecolor{oldgray}{HTML}{8A8A8A}\definecolor{mypurple}{HTML}{8E24AA}
\ifshowedits
  \newenvironment{newtext}{\color{editblue}}{}            
  \newenvironment{mytext}{\color{mypurple}}{}             
  \newcommand{\old}[1]{\textcolor{oldgray}{\sout{#1}}}     
\else
  \excludecomment{oldtext}
  \newcommand{\old}[1]{}
\fi

\newcommand{\ours}{\textsc{Wavefront}}
\newcommand{\K}{K}
\renewcommand{\P}{P}   

\mlsystitlerunning{Wavefront: Low-Rank Drafting with Parallel Causal Refinement}

\begin{document}

\twocolumn[
\mlsystitle{Wavefront: Low-Rank Drafting with Parallel Causal Refinement \\ for Speculative Decoding}

\mlsyssetsymbol{intern}{*}
\begin{mlsysauthorlist}
\mlsysauthor{Huy Huu Nguyen}{uo,intern}
\mlsysauthor{Xin Huang}{amzn}
\end{mlsysauthorlist}
\mlsysaffiliation{uo}{University of Oregon}
\mlsysaffiliation{amzn}{Amazon}
\mlsyscorrespondingauthor{Xin Huang}{xinxh@amazon.com, hxgy610@gmail.com}
\mlsyskeywords{speculative decoding, parallel drafting, Jacobi decoding, low-rank, vLLM}

\vskip 0.3in
]

\printAffiliationsAndNotice{\textsuperscript{*}Work done during an internship at Amazon. }

\begin{abstract}
Speculative decoding accelerates LLM inference by letting a cheap drafter propose tokens that the target model
verifies in parallel. Parallel block drafters such as DFlash and P-EAGLE propose a whole block in one forward
pass, but they lose the dependence between tokens in the block; DSpark restores part of it with a lightweight
correction head that runs sequentially. Measuring DSpark, we find that this head overrides the base prediction
on most slots, so the expensive base readout through the target's full lm\_head mostly provides a seed. We therefore propose \ours{}, which
follows a simple principle: \emph{shrink the base and strengthen the correction head}. \ours{} replaces the full lm\_head  with a learned low-rank base, conditions the correction head on the whole drafted prefix within the same block,
and resolves it with a few parallel Jacobi passes. 
A fixed pass budget makes the draft step free of data-dependent control flow, so we capture it as a single CUDA graph in vLLM with fused kernels.
On Qwen3-8B, \ours{} improves throughput over xPress, DSpark and
DFlash by 2.6\%, 10.7\% and 23.6\% on average across math, code, chat and agentic benchmarks, and by 9.6\%,
14.3\% and 28.9\% on SPEED-Bench long-context prompts. On Qwen3.8-27B it leads DFlash2, DSpark and xPress by
7.6\%, 10.6\% and 5.0\%, and by 7.4\%, 5.8\% and 6.0\% on the same long-context prompts. The gains persist as
serving load grows to concurrency 32.
\end{abstract}


\begin{figure}[!tb]
\centering
\includegraphics[width=0.9\columnwidth]{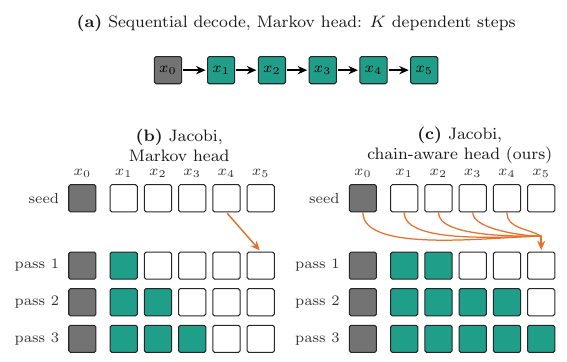}
\caption{Whole-prefix conditioning lets several slots settle per Jacobi pass (illustrative, $\K{=}5$).
(a) Sequential decode with the Markov head: one slot per step. (b) Jacobi with the Markov head: slot $k$ reads only
slot $k{-}1$, so a correction advances one slot per pass. (c) Jacobi with the chain-aware head: slot $k$ reads all earlier
slots in the block, so several settle per pass. Rows are passes; teal: settled; grey: anchor $x_0$; orange: what slot 5 reads
to compute pass 1. Measured settle times are in Table~\ref{tab:settle}.}
\label{fig:jacobi}
\end{figure}

\section{Introduction}
Speculative decoding \cite{leviathan2023,chen2023} accelerates LLM inference by amortizing an expensive target-model forward
pass over several tokens proposed by a cheaper drafter. Strong drafters such as EAGLE \cite{eagle,eagle2,eagle3} still extend
the draft one token at a time, so drafter latency grows with the number of proposed tokens. Parallel block drafters such as
DFlash \cite{dflash2026} and P-EAGLE \cite{peagle2026} remove this loop by proposing a block of $\K$ tokens in a single forward
pass. The price is intra-block dependence: each position is predicted without observing the tokens drafted before it, so
individually plausible predictions can combine into a sequence that the target rejects early.

DSpark \cite{dspark2026} restores part of this dependence with a lightweight Markov correction head: starting from the block's
base logits, it adds a low-rank bias indexed by the previously drafted token and decodes the block left to right. This raises a
systems question: \emph{which component actually determines the final draft?} If the head only nudges an already strong base,
spending most of the readout bandwidth on the base is sensible; if the head frequently decides the draft, that allocation is
backwards. Measuring DSpark directly, we find the latter. On GSM8K and HumanEval, its Markov head overrides 61--64\% of the
base predictions, and nearly half of the accepted tokens come from slots it overrides (\S\ref{sec:work}). These overrides sit on
top of base logits that cost a full read of the language-model head (lm\_head) at every draft step, the largest weight matrix
the drafter reads. This suggests a simple principle: \textbf{shrink the base and strengthen the correction head}.

The first challenge is \emph{strengthening the correction head}. DSpark already shows that such a correction is valuable: its
Markov head frequently overturns the base prediction and improves draft quality over the parallel backbone. The benefit, however, is limited
in two ways. In latency, each correction depends on the previously drafted token, so the head decodes the block left to right and
gives up part of the parallelism that made block drafting attractive. In quality, each correction observes only its previous
token, which bounds what the head can model. A stronger head must therefore predict better without decoding the block
sequentially. Jacobi iteration \cite{song2021,santilli2023} keeps the head parallel by updating all slots at once from the
previous iterate (Fig.~\ref{fig:jacobi}b), but the Markov head is a poor fit: its previous-token dependency moves a correction
forward by only one slot per pass, so refinement converges slowly (\S\ref{sec:conv}). We therefore replace it with a lightweight
causal attention head in which each slot attends to all earlier slots in the same draft block; we call it the \emph{chain-aware}
head. It addresses both limits at once: each correction draws on every earlier token, and a corrected token can inform every
later slot within one pass (Fig.~\ref{fig:jacobi}c). On GSM8K, it reaches 97\% of its converged acceptance length within five
passes, whereas the Markov head needs ten passes to reach 96\% of its own. 
xPress \cite{wang2026xpress}, developed concurrently, likewise corrects the block with a whole-prefix causal
head resolved by parallel Jacobi passes. The two designs differ in where they spend capacity. xPress keeps
the target's full lm\_head as its base and adds a correction on top, so every draft step still pays the full
vocabulary readout; \ours{} instead shrinks the base to a low-rank seed and moves that capacity into the
correction head. Additionally, their heads mix the prefix differently: xPress combines earlier slots with learned
weights that depend only on their positions, whereas our attention weights each earlier slot by its content,
so which slots matter can change with the drafted tokens.

The second challenge is \emph{shrinking the base}. Once the correction head carries the modeling work, the base only needs to
seed refinement, so \ours{} replaces the full lm\_head with a learned low-rank base that reads an eighth of its weights on
Qwen3-8B. This base gives the same seed as the full lm\_head (\S\ref{sec:seed}), and it also sets \ours{} apart from xPress, which
refines on top of the full lm\_head. In summary, we make the following contributions:
\begin{itemize}\setlength{\itemsep}{1pt}
\item \textbf{An analysis of where drafter capacity goes.}  We show that DSpark's lightweight Markov head, rather than its costlier full-vocabulary base, frequently decides the final draft, and that its previous-token dependency makes parallel refinement converge slowly
(\S\ref{sec:analysis}).
\item  We propose \textbf{\ours{}}, following the principle \emph{shrink the base and strengthen the
correction head}. A chain-aware causal head carries the modeling work and settles in a small, fixed number of
parallel Jacobi passes; a low-rank base supplies the seed, and a base-anchor objective keeps it informative.
Because the pass count is fixed, the whole draft step, refinement included, runs as a single CUDA graph in vLLM,
with fused kernels that score each slot without materializing the corrected full-vocabulary logits
(\S\ref{sec:method}, \S\ref{sec:inference}).
\item \textbf{Consistent end-to-end gains.} On Qwen3-8B, \ours{} improves throughput over xPress, DSpark and
DFlash by 2.6\%, 10.7\% and 23.6\% on average, and by 9.6\%, 14.3\% and 28.9\% on SPEED-Bench long-context
prompts; on Qwen3.8-27B it leads DFlash2, DSpark and xPress by 7.6\%, 10.6\% and 5.0\%. The gains persist as
serving load grows to concurrency 32 (\S\ref{sec:results}).
\end{itemize}

\section{Background}\label{sec:bg}
\paragraph{Speculative decoding.} A drafter proposes several future tokens, which the target verifies together in a single
forward pass; the target accepts the longest valid prefix, adds one token of its own, and continues from the verified sequence
\cite{leviathan2023,chen2023}. Let $T_{\text{draft}}$ and $T_{\text{verify}}$ denote the time to draft one block and to verify it,
and acceptance length (AL) the average number of tokens emitted per verification. The per-token latency is
\begin{equation}
  L = \frac{T_{\text{draft}} + T_{\text{verify}}}{\text{AL}},
\end{equation}
so a drafter reduces latency either by raising AL or by lowering $T_{\text{draft}}$, and the two often trade off.

\paragraph{Parallel block drafters.} Autoregressive drafters such as EAGLE \cite{eagle,eagle2,eagle3} spend one forward pass per
drafted token, so $T_{\text{draft}}$ grows with the draft length. DFlash \cite{dflash2026} and P-EAGLE \cite{peagle2026} remove
this sequential loop by predicting all $\K$ future positions, or \emph{slots}, in a single drafter forward pass conditioned on
features from the target model. Let $x_k$ denote the token drafted at slot $k$ and $x_0$ the last verified token, or anchor. We
call $x_{<k}=(x_0,\dots,x_{k-1})$ the \emph{drafted prefix} of slot $k$. Parallel drafters share the property that matters here:
slot $k$ is conditioned on the context and the anchor but not on the tokens $x_1,\dots,x_{k-1}$, so each slot is
a marginal prediction. The resulting base logits at slot $k$, $U_k\in\mathbb{R}^{V}$, are read out through the target's full
lm\_head, where $V$ is the vocabulary size; we call this readout the \emph{base}.

\paragraph{The Markov head (DSpark).} To restore dependence among slots, DSpark \cite{dspark2026} keeps the DFlash backbone and
adds a correction bias keyed on the previously drafted token, so that the logits at slot $k$ become
$\ell_k = U_k + W_2 W_1[x_{k-1}]$, where $W_1\in\mathbb{R}^{V\times r}$ is a token-code table ($W_1[x]$ is the row of token $x$),
$W_2\in\mathbb{R}^{V\times r}$ is a vocabulary readout, and $r{=}256$. Because the bias at slot $k$ depends only on the token at
slot $k{-}1$, the head is a first-order Markov model over the drafted block, which gives it its name. Decoding therefore proceeds
left to right: sample $x_k$ from $\ell_k$, look up $W_1[x_k]$, and correct slot $k{+}1$. Since the corrected logits define a
softmax distribution at every slot, exact draft probabilities remain available for lossless verification.

\paragraph{Jacobi iteration.} A causal chain over the $\K$ slots, $x_k = f_k(x_{<k})$ with $x_0$ fixed, can also be solved in
parallel as a fixed-point problem, an idea known as Jacobi decoding \cite{song2021,santilli2023}. Starting from a guess $x^{(0)}$,
the \emph{seed}, Jacobi iteration updates all slots at once,
\begin{equation}
  x_k^{(t+1)} = f_k\big(x_{<k}^{(t)}\big), \qquad k=1,\dots,\K.
\end{equation}
Because slot $k$ depends only on earlier slots, it becomes final after at most $k$ passes, so $\K$ passes always suffice and the
result equals the sequential greedy decode \cite{song2021}. The Markov head can therefore run as Jacobi iteration rather than a
left-to-right loop (Fig.~\ref{fig:jacobi}a--b); \S\ref{sec:conv} examines how many passes it needs in practice.


\begin{figure}[!tb]
\centering
\includegraphics[width=0.9\columnwidth]{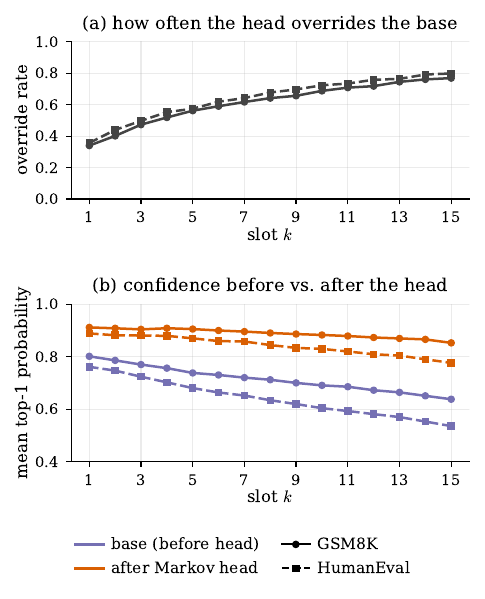}
\caption{The Markov head increasingly overrides the base deeper in the block (DSpark on its full-rank base, final checkpoint,
greedy, $\K{=}15$). (a) Override rate $P(C)$ per slot: fraction of slots where the head's token differs from the base's top-1
token. (b) Mean top-1 probability per slot of the base and of the distribution after the Markov head (solid: GSM8K; dashed:
HumanEval): base confidence decays with depth, while the corrected distribution stays more confident.}
\label{fig:flip}
\end{figure}

\section{Diagnosis: where should drafter capacity go?}\label{sec:analysis}
We start from DSpark, which pairs a parallel DFlash backbone with a sequential Markov head, and measure where its capacity is
spent. This section uses Qwen3-8B and a block-15 DFlash backbone trained with DSpark's recipe (\S\ref{sec:setup}).

\subsection{The correction head often determines the final draft}\label{sec:work}

\begin{table}[!tb]
\caption{How often DSpark's Markov head overrides its full-rank base, and whether the overrides survive verification. Rates are
pooled over all $\K{=}15$ slots, except $P_1(A\mid C)$, measured at the first slot.}
\label{tab:flip}
\vspace{5pt}
\centering\footnotesize\setlength{\tabcolsep}{3pt}
\begin{tabular}{lcccc}
\toprule
dataset & $P(C)$ & $P(C\mid A)$ & $P_1(A\mid C)$ & $P(C\mid p^{\mathrm{base}}\!\ge\!0.9)$ \\
\midrule
GSM8K     & 61\% & 48\% & 87\% & 36\% \\
HumanEval & 64\% & 45\% & 83\% & 31\% \\
\bottomrule
\end{tabular}
\end{table}

DSpark reads its base logits from the target's frozen lm\_head, a $4096{\times}V$ projection of about 1.2\,GB in bf16 for
Qwen3-8B, whereas the Markov head that corrects it consists of two $V{\times}256$ tables of 78\,MB each. This allocation is
sensible only if the base produces most final predictions. For every drafted slot, let $C$ denote the event that the head
overrides the base argmax, and $A$ the event that the slot is accepted by the target. Table~\ref{tab:flip} reports $P(C)$; the
share of accepted tokens that come from overrides, $P(C\mid A)$; how often overrides are accepted at the first slot, whose
acceptance does not depend on earlier drafted tokens, $P_1(A\mid C)$; and how often the head overrides a confident base,
$P(C\mid p^{\mathrm{base}}\ge 0.9)$, on GSM8K (300 prompts) and HumanEval (164 prompts) under greedy decoding.

The Markov head overrides the base argmax on 61--64\% of drafted slots, rising from 34--36\% at the first slot to 77--80\% at the
deepest slots (Fig.~\ref{fig:flip}a). 45--48\% of all accepted tokens come from overridden slots, and the overrides are usually
right: 83--87\% of them are accepted at the first slot. The head even overrides 31--36\% of the slots where the base assigns at
least 90\% probability to its top token. Overrides grow with depth because the base cannot see the tokens drafted before each
slot, so its confidence falls along the block, while the head stays more confident at every slot (Fig.~\ref{fig:flip}b). The
full-vocabulary readout thus mostly provides a seed that the much smaller head rewrites. Refinement still needs that seed, but
paying full lm\_head bandwidth for it spends the drafter's capacity in the wrong place. This motivates our principle,
\textbf{shrink the base and strengthen the correction head}, whose two halves we examine next.

\subsection{A stronger correction head stays fast by reading the whole prefix in parallel}\label{sec:conv}

\begin{table}[!tb]
\caption{Passes until the block stops changing ($\K{=}15$, $\P{=}15$, greedy; same runs as Fig.~\ref{fig:conv}): mean, median, and
fraction of blocks settled within 5 and 10 passes.}
\label{tab:settle}
\vspace{5pt}
\centering\footnotesize\setlength{\tabcolsep}{3pt}
\begin{tabular}{llcccc}
\toprule
head & dataset & mean & med. & $\le$5 & $\le$10 \\
\midrule
Markov & GSM8K       & 10.7 & 11 & 9\%  & 42\% \\
chain-aware & GSM8K  & 5.7  & 6  & 49\% & 80\% \\
Markov & HumanEval   & 11.2 & 12 & 8\%  & 35\% \\
chain-aware & HumanEval & 7.0 & 8  & 38\% & 72\% \\
\bottomrule
\end{tabular}
\end{table}


The first half of the principle asks for a head that predicts better without adding drafting latency. DSpark's Markov head
is limited on both fronts: each correction observes only the previous drafted token, and its latency is low only because each
correction is tiny, since the head still corrects the slots one at a time, in $\K$ sequential steps; a more expressive correction
would pay its cost $\K$ times. A stronger head therefore needs a richer view of the drafted prefix and a way to correct all slots
in parallel. Jacobi iteration (\S\ref{sec:bg}) offers the latter, and the Markov head can use it directly: its base logits are
fixed per block and each greedy correction depends only on the previous slot, so we seed every slot with the base argmax,
correct all slots from the previous pass, and repeat, which reproduces the sequential decode exactly within $\K$ passes. The
problem is speed, not correctness: a correction at slot $k$ reaches slot $k{+}1$ only in the next pass (Fig.~\ref{fig:jacobi}b).

\begin{figure}[!tb]
\centering
\IfFileExists{figures/fig_convergence_xin_1030.pdf}{\includegraphics[width=0.9\columnwidth]{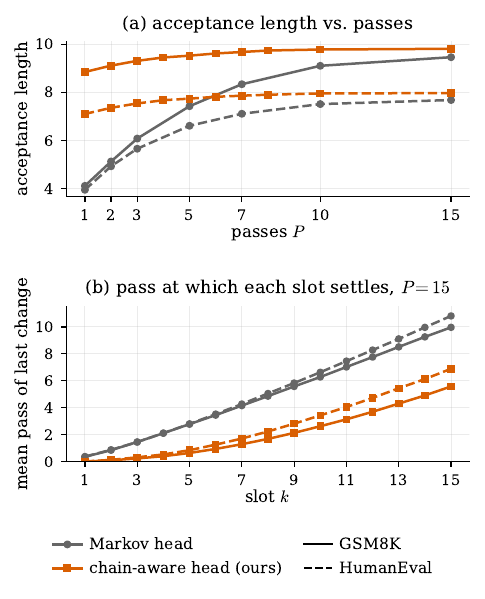}}{\todo{fig\_convergence.pdf: run the sweep (bench\_conv) and regenerate}}
\caption{The chain-aware head converges in far fewer Jacobi passes than the Markov head ($\K{=}15$, greedy). (a) Acceptance length
vs.\ passes $\P$. (b) Mean pass at which each slot's token last changes ($\P{=}15$).}
\label{fig:conv}
\end{figure}


\textbf{Markov head: slow convergence.} On GSM8K, the Markov head's acceptance length is only 4.11 after one
pass, reaches 9.11 after ten, and matches the sequential 9.46 only at fifteen (Fig.~\ref{fig:conv}a); its blocks stop changing
after 10.7 passes on average, and only 9\% settle within five (Table~\ref{tab:settle}).

\textbf{Chain-aware head: whole-prefix reads.} Our chain-aware head (\S\ref{sec:method}) lets slot $k$ read
compact codes for its whole drafted prefix through small causal attention over the block's slots, so a pass stays cheap and a
correction at one slot reaches every later slot in the next pass. On GSM8K, it reaches 90\% of its converged acceptance length
after one pass (8.85 of 9.81) and 97\% after five; its blocks stop changing after 5.7 passes on average, and 49\% settle within
five. It also converges to a better draft: 0.35 more accepted tokens on GSM8K (9.81 vs.\ 9.46) and 0.29 on HumanEval
(7.97 vs.\ 7.68).

\subsection{A low-rank base is enough to seed refinement}\label{sec:seed}
The second half of the principle asks for a base that is cheaper but still gives the head a good seed, the block of base
predictions from which refinement begins. We train three drafters that are identical except for the base: the target's full
lm\_head, and two learned low-rank bases of rank 512 and 256 (\S\ref{sec:method}). Before any refinement pass, the three bases
produce the same seed quality: on GSM8K, the base's top token matches the target's 96.2--96.3\% of the time at slot 1 and
47.5--48.8\% at slot 15 for all three, and at no reported slot of GSM8K, HumanEval or MT-Bench do they differ by more than 1.3
points (Table~\ref{tab:slotacc} in Appendix~\ref{app:diag}). A full-vocabulary readout is therefore not needed to seed refinement, and we replace it with a
learned low-rank base; \S\ref{sec:abl} evaluates its end-to-end effect.

\section{\ours{}}\label{sec:method}

\begin{figure*}[!tb]
\centering
\includegraphics[width=0.8\textwidth]{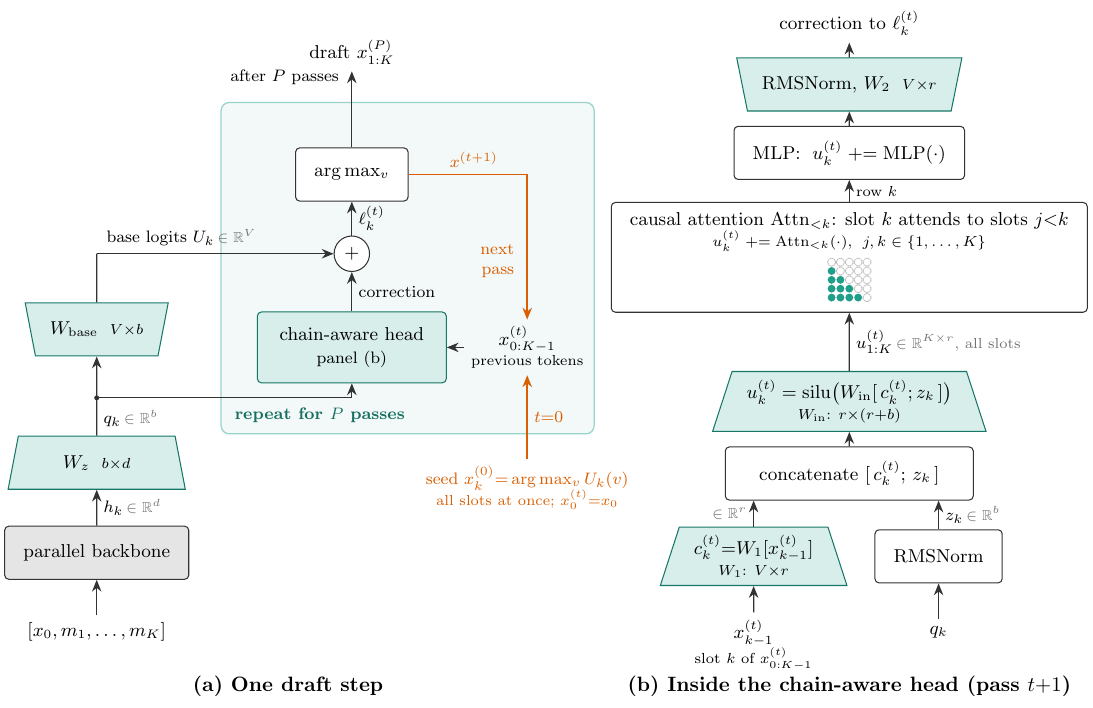}
\caption{\ours{} on a parallel block drafter (e.g., DFlash or P-EAGLE). (a) One draft step: the
backbone and the low-rank base run once, giving the base logits $U_k$ and the seed $x^{(0)}$; the
chain-aware head then refines the block for $P$ passes, each pass reading the previous block $x^{(t)}$.
(b) One pass of the chain-aware head for slot $k$: the code of the predecessor token is concatenated
with the normalized feature $z_k$, mixed across earlier slots by causal attention, and projected to a
correction of the base logits. Teal: weights learned by \ours{}.}
\label{fig:arch}
\end{figure*}

\ours{} realizes the two halves of the principle from \S\ref{sec:analysis}. A cheap low-rank base provides the seed
(\S\ref{sec:lowrank}), and the chain-aware head corrects it using the whole drafted prefix within the same draft block, refined over a few parallel passes
(\S\ref{sec:chain-head}). The parallel backbone itself is unchanged. 
Figure~\ref{fig:arch} shows the architecture, and Appendix~\ref{app:notation} summarizes the notation.

\subsection{A low-rank base for the seed}\label{sec:lowrank}
The drafter receives DSpark's anchor-first block $[x_0, m_1, \ldots, m_\K]$, where $x_0$ is the last verified token (the anchor)
and the mask token $m_k$ marks slot $k$. Let $h_k\in\mathbb{R}^{d}$, $k=1,\ldots,\K$, denote the unchanged backbone hidden state from which
slot $k$ is predicted.
Instead of applying the target's full $d{\times}V$ lm\_head, \ours{} projects the hidden state once,
\begin{equation}
  q_k = W_z h_k, \qquad U_k = W_{\mathrm{base}} q_k,
\end{equation}
where $W_z\in\mathbb{R}^{b\times d}$ and $W_{\mathrm{base}}\in\mathbb{R}^{V\times b}$ with $b\ll d$. The readout then reads $V\times b$
weights instead of the $V \times d$ of the full lm\_head, a $d/b$ reduction; the ranks used for each target are given in
\S\ref{sec:setup}. The same $q_k$ is reused by the chain-aware causal head, so the projection is computed only once. The goal of $U_k$ is not
to be a standalone language-model distribution; it is to provide an informative seed for refinement. 

\subsection{A chain-aware causal head}\label{sec:chain-head}
A previous-token Markov head is cheap but places all causal information on a one-hop chain. \ours{} instead lets every slot read
compact codes for its entire drafted prefix. 
Let $x^{(t)}=(x_0^{(t)},x_1^{(t)},\ldots,x_\K^{(t)})$ denote the block after $t$ refinement passes,
including the anchor. We seed the block from the low-rank base and keep the anchor fixed:
\begin{equation}\label{eq:refine-seed}
 x_k^{(0)} = \arg\max_v U_k(v), \qquad x_0^{(t)}=x_0.
\end{equation}
where $U_k(v)$ is the base logit of vocabulary token $v$.
Pass $t{+}1$ reads $x^{(t)}$ and writes $x^{(t+1)}$, so the first pass ($t{=}0$) reads the seed and the final draft is $x^{(P)}$.
In pass $t{+}1$, slot $k$ receives the code of its predecessor token in the current block,
\begin{equation}\label{eq:refine-code}
 c_k^{(t)} = W_1[x_{k-1}^{(t)}]\in\mathbb{R}^{r}.
\end{equation}
$\mathrm{Attn}_{<k}$ is multi-head causal attention over the $\K$ slots (configuration in
\S\ref{sec:training}); it operates only on the $r$-dimensional block latents, not on the full
context, so its cost is independent of prompt length.
We reuse the names $W_1\in\mathbb{R}^{V\times r}$ and $W_2\in\mathbb{R}^{V\times r}$ (the readout below) for \ours{}'s own tables:
they have the same shapes as DSpark's Markov tables ($r{=}256$) but are trained as part of the head. Every slot therefore reads the block from the previous pass, so all slots can be updated in
parallel. With $z_k=\mathrm{RMSNorm}(q_k)$ fixed across passes, one head evaluation computes 
\begin{align}
 u_k^{(t)} &= \mathrm{silu}\!\left(W_{\mathrm{in}}[c_k^{(t)};z_k]\right), \\
 u_k^{(t)} &\mathrel{+}= \mathrm{Attn}_{<k}\!\left(\mathrm{RMSNorm}(u^{(t)})\right)_k, \\
 u_k^{(t)} &\mathrel{+}= \mathrm{MLP}\!\left(\mathrm{RMSNorm}(u_k^{(t)})\right), \\
 \ell_k^{(t)} &= U_k + W_2\,\mathrm{RMSNorm}(u_k^{(t)}).
\end{align}
The corrected logits produce the next block at every slot simultaneously,
\begin{equation}\label{eq:refine-update}
 x_k^{(t+1)} = \arg\max_v \ell_k^{(t)}(v).
\end{equation}
$\mathrm{Attn}_{<k}$ is strictly causal over the $\K$ slots. The vocabulary tables $W_1$ and $W_2$ dominate the head size; the
attention and MLP add comparatively little parameter or compute cost.

Together, Eqs.~\ref{eq:refine-code}--\ref{eq:refine-update} define one refinement pass. A Markov correction for slot $k$
observes only $x_{k-1}^{(t)}$, whereas ours depends on $x_{<k}^{(t)}$, which both improves the converged draft and lets one
pass carry information across many slots (\S\ref{sec:conv}).


\subsection{Training the base and head together}\label{sec:train}
We follow DSpark's training recipe and add one term. DSpark's loss on the logits $s$ at slot $k$ is
$\mathcal{L}(s)=0.1\,\mathrm{CE}\big(\mathrm{softmax}(s),\,y_k\big)+0.9\,\big\|\mathrm{softmax}(s)-p_k\big\|_1$,
where $y_k$ is the target's token and $p_k$ its distribution. We apply it to the corrected logits, as
DSpark does, and add the same loss on the base logits, a \emph{base anchor}:
\begin{equation}\label{eq:loss}
  \mathcal{L}_{\text{total}} \;=\; \sum_{k=1}^{\K} e^{-(k-1)/4}\,\Big[\,\underbrace{\mathcal{L}(\ell_k)}_{\text{corrected loss}}
  \;+\; \alpha\,\underbrace{\mathcal{L}(U_k)}_{\text{base anchor}}\,\Big].
\end{equation}
Without the anchor, the stronger head can absorb the prediction task and leave the low-rank base a
poor seed (\S\ref{sec:seed}); with it, the base is trained to predict the target on its own. The weight
$\alpha$ is annealed linearly from 1.0 to 0.3 over training: the larger initial weight lets the base learn
a useful prediction quickly, while the remaining weight keeps it a good seed throughout training. The
factor $e^{-(k-1)/4}$, also from DSpark, weights early slots more, since a slot can only be accepted if every
slot before it is.


\subsection{Inference implementation}\label{sec:inference}
At inference, we run $P$ refinement passes of \S\ref{sec:chain-head}, starting from the seed in Eq.~\ref{eq:refine-seed}. The
backbone features and base logits are computed once per draft block and reused in every pass. Strict causality guarantees that the
block settles within $\K$ passes, but whole-prefix conditioning settles most of it in a few (\S\ref{sec:conv}), so we use $P<\K$;
the values used in our experiments are given in \S\ref{sec:setup}, and \S\ref{sec:passes} sweeps $P$. With greedy decoding, each pass takes the argmax of the corrected logits and verification reduces to exact greedy matching. With sampling, each drafted token is drawn from its slot's corrected distribution, and verification uses the distribution from which it was actually drawn, so standard rejection sampling remains lossless.
Keeping $P$ fixed, rather than stopping at convergence, avoids host synchronization,
so the whole draft step runs as a single CUDA graph. Appendix~\ref{app:impl} describes our vLLM implementation and the fused kernels that keep refinement cheap at serving time.



\section{Experimental setup}\label{sec:setup}
\subsection{Models and training}\label{sec:training}
We train \ours{} for two targets, Qwen3-8B and Qwen3.8-27B. Training prompts are sampled from Open-PerfectBlend \cite{xu2024perfect},
UltraChat \cite{ding2023ultrachat} and code instructions (25K prompts each from Magicoder \cite{wei2024magicoder} and OpenCodeInstruct
\cite{ahmad2025opencodeinstruct}), with responses regenerated by each target. Block sizes follow the released drafters: 15 draft
tokens on Qwen3-8B (as DFlash, DSpark and xPress) and 7 on Qwen3.8-27B (as DFlash2); a checkpoint serves any speculation length up
to its block size.

For both targets, we train a five-layer DFlash backbone from scratch for 10 epochs at global batch 512, with learning rate
$6\times10^{-4}$, cosine decay and 4\% warmup, following DSpark's schedule \cite{dspark2026}. The two targets differ only in the
target layers the backbone reads ($\{1,9,17,25,33\}$ on Qwen3-8B and $\{5,19,33,47,61\}$ on Qwen3.8-27B) and in the rank of the
base ($b{=}512$ and $b{=}1024$,  scaled with the size of target). On both, the chain-aware head uses $r{=}256$ and multi-head causal attention with 8 heads of width 64
and slot-position RoPE, and runs $P{=}5$ refinement passes. Appendix~\ref{app:peagle} gives the setup of the P-EAGLE backbone.

\subsection{Baselines}\label{sec:baselines}
We compare against the strongest released parallel drafters for each target.

\textbf{DFlash and DFlash2} \cite{dflash2026,inco2026dflash2}. DFlash drafts a block in parallel from target hidden states, and
DFlash2 adds two-tap dynamic convolutions to its trunk and selects a path through the top-16 candidates at each slot with a
low-rank pairwise scorer. On Qwen3-8B we use the official DFlash release. DFlash2 has no official Qwen3-8B release, so we report a
seven-token DFlash2 trained by Red Hat \cite{speculators2025} for reference only. On Qwen3.8-27B we use the official DFlash2 release.

\textbf{DSpark} \cite{dspark2026} corrects the DFlash block with a sequential previous-token Markov head. On Qwen3-8B we use the
public 15-token Markov-head checkpoint released with xPress \cite{wang2026xpress}; on Qwen3.8-27B we use a public release.

\textbf{xPress} \cite{wang2026xpress} refines the DFlash block over six parallel passes with a causal mixer whose weights are
learned per position. On Qwen3-8B we use the official release with its vLLM implementation. Qwen3.8-27B has no official release,
so we train xPress on the released DFlash2 drafter following its official recipe.

All baselines are served as released, except that adaptive verification is disabled for every drafter.\footnote{DSpark uses its
confidence head for adaptive verification. A logit-based estimator (vLLM PR \#52228) gives the same capability to drafters without
such a head, so we disable it for all drafters to compare drafting alone.} We do not include Domino \cite{huang2026domino}, whose
GRU correction is sequential across the block, the same class as DSpark's Markov head.

\paragraph{Controlled ablations.} \S\ref{sec:abl}--\S\ref{sec:cost} isolate our design on Qwen3-8B. Every variant shares the
DFlash backbone, training data and schedule, and differs only in the head or the base. In this setting, \emph{DSpark} is its Markov
head retrained with our backbone, and \emph{full-rank} keeps \ours{}'s chain-aware head but reads the base logits from the full
lm\_head.

\subsection{Evaluation datasets}
Following DSpark, we use the twelve sets of the DeepSpec evaluation suite \cite{dspark2026} in four domains: math, GSM8K \cite{cobbe2021gsm8k}, MATH-500
\cite{lightman2024verify} and AIME 2024/2025 \cite{aime2024,aime2025}; code, HumanEval \cite{chen2021codex}, MBPP \cite{austin2021program},
LBPP \cite{matton2024leakage} and LiveCodeBench \cite{jain2025livecodebench}; chat, MT-Bench \cite{zheng2023judging}, Alpaca
\cite{taori2023alpaca} and Arena-Hard v2 \cite{li2025arenahard}; and agent, SWE-bench \cite{jimenez2024swebench}. For long inputs we use
the throughput split of SPEED-Bench \cite{abramovich2026speed}: its low-entropy (\emph{code \& sorting}) and high-entropy (\emph{books \& writing}) prompts at 2K, 8K,
16K and 32K input tokens.


\subsection{Measurement protocol}\label{sec:protocol}
All drafters are served in vLLM v0.32
nightly build (that merged xPress) with greedy decoding ($T{=}0$). Qwen3-8B runs without thinking, as in DFlash and DSpark
\cite{dflash2026,dspark2026}, on one GPU: an H200 for the concurrency-1 results of Table~\ref{tab:main} and an H100 for all other
experiments.   Qwen3.8-27B runs with thinking enabled, as in DFlash2 \cite{inco2026dflash2}, on two H100 GPUs with tensor parallelism 2. Within each comparison, all drafters share the same hardware, and speedups are over autoregressive decoding (AR) of the same target
on that hardware. We report \emph{acceptance length} (AL), $1+\text{accepted tokens}/\text{steps}$, and \emph{throughput} (TPS),
generated tokens per second over all concurrent requests. Appendix~\ref{app:serving} gives prompt counts, vLLM builds and other
details.

\section{Results}\label{sec:results}
We organize the evaluation around four questions implied by the design: (1) does the reallocation improve end-to-end speculative
decoding across targets and backbones (\S\ref{sec:headline}), (2) how does the throughput advantage change with serving concurrency
(\S\ref{sec:conc}), (3) is the low-rank base actually sufficient under the same chain-aware head (\S\ref{sec:abl}), and (4) how many
refinement passes are needed, and where does the saved time come from (\S\ref{sec:passes}--\S\ref{sec:cost})? All comparisons use
the protocol of \S\ref{sec:setup}.

\begin{table*}[tbp]
\caption{AL, speedup and TPS gain on Qwen3-8B and Qwen3.8-27B at concurrency 1, $T{=}0$. AR column: its TPS. Cells: AL / speedup over AR, best speedup per row in bold. TPS gain: \ours{} over each baseline; Average rows compare mean TPS. \ours{}: rank-512 (8B) and rank-1024 (27B) base, $P{=}5$. $^\dagger$DFlash2 trained by Red Hat \cite{speculators2025}, 7 draft tokens; reference only.}
\label{tab:main}
\vspace{5pt}
\centering\scriptsize\setlength{\tabcolsep}{3pt}
\resizebox{\textwidth}{!}{%
\begin{tabular}{@{}lrccccccccrccccccc@{}}
\toprule
 & \multicolumn{9}{c}{Qwen3-8B, speculation length 15} & \multicolumn{8}{c}{Qwen3.8-27B, speculation length 7} \\
\cmidrule(lr){2-10}\cmidrule(lr){11-18}
 & AR & \multicolumn{5}{c}{AL / speedup over AR ($\times$)} & \multicolumn{3}{c}{TPS gain (\%) vs} & AR & \multicolumn{4}{c}{AL / speedup over AR ($\times$)} & \multicolumn{3}{c}{TPS gain (\%) vs} \\
\cmidrule(lr){3-7}\cmidrule(lr){8-10}\cmidrule(lr){12-15}\cmidrule(lr){16-18}
 & TPS & DFlash & \textcolor{gray}{DFlash2$^\dagger$} & DSpark & xPress & \ours{} & xPress & DSpark & DFlash & TPS & DFlash2 & DSpark & xPress & \ours{} & xPress & DSpark & DFlash2 \\
\midrule
\multicolumn{18}{@{}l}{\textbf{DeepSpec Evaluation Suite}} \\
\multicolumn{18}{@{}l}{Math} \\
\hspace{3pt}GSM8K & 194 & 6.2/4.87 & \textcolor{gray}{5.9/4.46} & 9.6/6.71 & 9.9/7.25 & 9.6/\textbf{7.42} & \textcolor{gaingreen}{2.3} & \textcolor{gaingreen}{10.5} & \textcolor{gaingreen}{52.3} & 84 & 5.0/3.69 & 5.1/3.70 & 5.3/3.95 & 5.4/\textbf{4.10} & \textcolor{gaingreen}{3.8} & \textcolor{gaingreen}{10.8} & \textcolor{gaingreen}{11.1} \\
\hspace{3pt}MATH-500 & 194 & 7.8/6.19 & \textcolor{gray}{5.8/4.47} & 9.3/6.78 & 9.7/7.31 & 9.3/\textbf{7.45} & \textcolor{gaingreen}{1.9} & \textcolor{gaingreen}{9.8} & \textcolor{gaingreen}{20.4} & 85 & 4.7/3.53 & 4.6/3.49 & 4.8/3.63 & 4.9/\textbf{3.83} & \textcolor{gaingreen}{5.5} & \textcolor{gaingreen}{9.7} & \textcolor{gaingreen}{8.5} \\
\hspace{3pt}AIME 2024 & 194 & 7.4/5.92 & \textcolor{gray}{5.5/4.26} & 8.3/6.06 & 8.6/6.58 & 8.2/\textbf{6.62} & \textcolor{gaingreen}{0.7} & \textcolor{gaingreen}{9.4} & \textcolor{gaingreen}{11.9} & 85 & 4.4/3.28 & 4.3/3.27 & 4.4/3.30 & 4.5/\textbf{3.50} & \textcolor{gaingreen}{6.1} & \textcolor{gaingreen}{7.0} & \textcolor{gaingreen}{6.7} \\
\hspace{3pt}AIME 2025 & 194 & 6.9/5.51 & \textcolor{gray}{5.2/4.04} & 7.9/5.76 & 8.1/6.19 & 7.8/\textbf{6.36} & \textcolor{gaingreen}{2.7} & \textcolor{gaingreen}{10.5} & \textcolor{gaingreen}{15.4} & 86 & 4.3/3.25 & 4.3/3.26 & 4.4/3.38 & 4.5/\textbf{3.50} & \textcolor{gaingreen}{3.6} & \textcolor{gaingreen}{7.4} & \textcolor{gaingreen}{7.7} \\
\multicolumn{18}{@{}l}{Code} \\
\hspace{3pt}HumanEval & 194 & 6.4/5.02 & \textcolor{gray}{4.9/3.80} & 7.7/5.56 & 8.0/6.01 & 7.8/\textbf{6.21} & \textcolor{gaingreen}{3.3} & \textcolor{gaingreen}{11.7} & \textcolor{gaingreen}{23.6} & 85 & 4.5/3.37 & 4.3/3.26 & 4.6/3.49 & 4.7/\textbf{3.68} & \textcolor{gaingreen}{5.4} & \textcolor{gaingreen}{12.9} & \textcolor{gaingreen}{9.2} \\
\hspace{3pt}MBPP & 195 & 5.7/4.45 & \textcolor{gray}{4.5/3.44} & 6.9/4.95 & 7.1/5.31 & 6.9/\textbf{5.43} & \textcolor{gaingreen}{2.2} & \textcolor{gaingreen}{9.6} & \textcolor{gaingreen}{22.1} & 87 & 4.5/3.29 & 4.2/3.17 & 4.4/3.22 & 4.5/\textbf{3.42} & \textcolor{gaingreen}{6.2} & \textcolor{gaingreen}{7.9} & \textcolor{gaingreen}{4.0} \\
\hspace{3pt}LBPP & 195 & 5.2/4.15 & \textcolor{gray}{4.5/3.44} & 6.2/4.53 & 6.5/4.93 & 6.4/\textbf{5.16} & \textcolor{gaingreen}{4.8} & \textcolor{gaingreen}{13.9} & \textcolor{gaingreen}{24.5} & 85 & 3.8/2.83 & 3.6/2.75 & 3.8/2.94 & 3.9/\textbf{3.03} & \textcolor{gaingreen}{3.1} & \textcolor{gaingreen}{10.2} & \textcolor{gaingreen}{7.1} \\
\hspace{3pt}LiveCodeBench & 192 & 7.0/5.29 & \textcolor{gray}{5.0/3.74} & 7.8/5.39 & 8.0/5.76 & 8.0/\textbf{6.11} & \textcolor{gaingreen}{6.1} & \textcolor{gaingreen}{13.4} & \textcolor{gaingreen}{15.5} & 86 & 3.5/2.64 & 3.5/2.63 & 3.5/2.69 & 3.6/\textbf{2.82} & \textcolor{gaingreen}{4.8} & \textcolor{gaingreen}{7.2} & \textcolor{gaingreen}{6.8} \\
\multicolumn{18}{@{}l}{Chat} \\
\hspace{3pt}MT-Bench & 195 & 3.2/2.57 & \textcolor{gray}{3.4/2.59} & 4.2/3.08 & 4.3/3.31 & 4.2/\textbf{3.37} & \textcolor{gaingreen}{1.9} & \textcolor{gaingreen}{9.5} & \textcolor{gaingreen}{31.1} & 85 & 3.8/2.87 & 3.6/2.70 & 3.8/2.90 & 4.0/\textbf{3.10} & \textcolor{gaingreen}{6.9} & \textcolor{gaingreen}{14.8} & \textcolor{gaingreen}{8.0} \\
\hspace{3pt}Alpaca & 195 & 2.9/2.31 & \textcolor{gray}{3.2/2.44} & 4.0/2.88 & 4.1/3.10 & 4.0/\textbf{3.21} & \textcolor{gaingreen}{3.3} & \textcolor{gaingreen}{11.2} & \textcolor{gaingreen}{38.9} & 84 & 3.7/2.75 & 3.5/2.55 & 3.8/2.85 & 4.0/\textbf{3.04} & \textcolor{gaingreen}{6.7} & \textcolor{gaingreen}{19.2} & \textcolor{gaingreen}{10.5} \\
\hspace{3pt}Arena-Hard v2 & 193 & 3.5/2.73 & \textcolor{gray}{3.6/2.75} & 4.1/2.95 & 4.2/\textbf{3.20} & 3.9/3.15 & \textcolor{lossred}{-1.8} & \textcolor{gaingreen}{6.8} & \textcolor{gaingreen}{15.4} & 86 & 3.7/2.77 & 3.5/2.66 & 3.7/2.78 & 3.7/\textbf{2.89} & \textcolor{gaingreen}{4.0} & \textcolor{gaingreen}{8.6} & \textcolor{gaingreen}{4.3} \\
\multicolumn{18}{@{}l}{Agent} \\
\hspace{3pt}SWE-bench & 193 & 3.6/2.89 & \textcolor{gray}{3.5/2.76} & 4.4/3.24 & 4.6/3.52 & 4.5/\textbf{3.61} & \textcolor{gaingreen}{2.5} & \textcolor{gaingreen}{11.2} & \textcolor{gaingreen}{25.0} & 86 & 4.5/3.42 & 4.3/3.23 & 4.6/3.49 & 4.7/\textbf{3.63} & \textcolor{gaingreen}{4.0} & \textcolor{gaingreen}{12.4} & \textcolor{gaingreen}{6.1} \\
\cmidrule(lr){1-18}
\textit{Average} & 191 & 5.5/4.32 & \textcolor{gray}{4.6/3.52} & 6.7/4.82 & 6.9/5.21 & 6.7/\textbf{5.34} & \textcolor{gaingreen}{2.6} & \textcolor{gaingreen}{10.7} & \textcolor{gaingreen}{23.6} & 85 & 4.2/3.14 & 4.1/3.06 & 4.3/3.22 & 4.4/\textbf{3.38} & \textcolor{gaingreen}{5.0} & \textcolor{gaingreen}{10.6} & \textcolor{gaingreen}{7.6} \\
\midrule
\multicolumn{18}{@{}l}{\textbf{SPEED-Bench long context}} \\
\hspace{3pt}Code \& sorting 2K & 190 & 4.2/3.19 & \textcolor{gray}{4.0/2.94} & 5.1/3.50 & 5.1/3.71 & 5.0/\textbf{3.82} & \textcolor{gaingreen}{3.0} & \textcolor{gaingreen}{9.2} & \textcolor{gaingreen}{19.6} & 86 & 4.1/3.04 & 4.1/3.07 & 4.1/3.09 & 4.4/\textbf{3.35} & \textcolor{gaingreen}{8.4} & \textcolor{gaingreen}{9.1} & \textcolor{gaingreen}{10.2} \\
\hspace{3pt}Code \& sorting 8K & 176 & 2.2/1.66 & \textcolor{gray}{2.6/1.89} & 2.7/1.89 & 2.7/1.95 & 3.1/\textbf{2.30} & \textcolor{gaingreen}{17.4} & \textcolor{gaingreen}{21.3} & \textcolor{gaingreen}{38.4} & 83 & 4.0/2.79 & 4.1/2.91 & 4.0/2.86 & 4.2/\textbf{3.00} & \textcolor{gaingreen}{4.9} & \textcolor{gaingreen}{3.1} & \textcolor{gaingreen}{7.5} \\
\hspace{3pt}Code \& sorting 16K & 159 & 1.7/1.29 & \textcolor{gray}{2.4/1.70} & 2.0/1.42 & 2.0/1.44 & 2.1/\textbf{1.58} & \textcolor{gaingreen}{9.6} & \textcolor{gaingreen}{11.6} & \textcolor{gaingreen}{22.4} & 79 & 3.9/2.58 & 4.1/2.69 & 3.9/2.61 & 4.1/\textbf{2.78} & \textcolor{gaingreen}{6.5} & \textcolor{gaingreen}{3.3} & \textcolor{gaingreen}{7.8} \\
\hspace{3pt}Code \& sorting 32K & 136 & 1.6/1.21 & \textcolor{gray}{2.6/1.74} & 2.0/1.41 & 2.0/1.47 & 2.1/\textbf{1.56} & \textcolor{gaingreen}{6.0} & \textcolor{gaingreen}{10.4} & \textcolor{gaingreen}{28.5} & 73 & 4.0/2.35 & 4.4/\textbf{2.52} & 4.0/2.35 & 4.2/2.48 & \textcolor{gaingreen}{5.5} & \textcolor{lossred}{-1.6} & \textcolor{gaingreen}{5.5} \\
\hspace{3pt}Books \& writing 2K & 191 & 2.3/1.82 & \textcolor{gray}{2.5/1.89} & 2.7/1.97 & 2.8/2.12 & 2.9/\textbf{2.29} & \textcolor{gaingreen}{8.1} & \textcolor{gaingreen}{16.2} & \textcolor{gaingreen}{26.2} & 87 & 3.0/2.24 & 2.9/2.17 & 3.0/2.26 & 3.1/\textbf{2.40} & \textcolor{gaingreen}{6.2} & \textcolor{gaingreen}{10.6} & \textcolor{gaingreen}{7.1} \\
\hspace{3pt}Books \& writing 8K & 177 & 2.7/2.06 & \textcolor{gray}{3.0/2.21} & 3.2/2.28 & 3.3/2.38 & 3.5/\textbf{2.62} & \textcolor{gaingreen}{10.0} & \textcolor{gaingreen}{14.9} & \textcolor{gaingreen}{27.1} & 83 & 3.0/2.15 & 2.9/2.09 & 3.0/2.19 & 3.1/\textbf{2.28} & \textcolor{gaingreen}{4.1} & \textcolor{gaingreen}{9.1} & \textcolor{gaingreen}{6.0} \\
\hspace{3pt}Books \& writing 16K & 161 & 2.0/1.52 & \textcolor{gray}{2.6/1.91} & 2.5/1.78 & 2.5/1.82 & 2.9/\textbf{2.14} & \textcolor{gaingreen}{17.4} & \textcolor{gaingreen}{19.9} & \textcolor{gaingreen}{40.4} & 80 & 3.0/2.08 & 2.9/2.05 & 3.0/2.13 & 3.1/\textbf{2.23} & \textcolor{gaingreen}{4.7} & \textcolor{gaingreen}{8.8} & \textcolor{gaingreen}{7.2} \\
\hspace{3pt}Books \& writing 32K & 131 & 1.6/1.27 & \textcolor{gray}{2.9/1.87} & 2.5/1.63 & 2.4/1.65 & 2.8/\textbf{1.86} & \textcolor{gaingreen}{13.0} & \textcolor{gaingreen}{14.0} & \textcolor{gaingreen}{47.0} & 74 & 2.9/1.91 & 2.9/1.94 & 2.9/1.90 & 3.1/\textbf{2.04} & \textcolor{gaingreen}{7.4} & \textcolor{gaingreen}{5.2} & \textcolor{gaingreen}{6.8} \\
\cmidrule(lr){1-18}
\textit{Average} & 165 & 2.3/1.75 & \textcolor{gray}{2.8/2.02} & 2.8/1.99 & 2.9/2.07 & 3.0/\textbf{2.27} & \textcolor{gaingreen}{9.6} & \textcolor{gaingreen}{14.3} & \textcolor{gaingreen}{28.9} & 81 & 3.5/2.39 & 3.5/2.43 & 3.5/2.42 & 3.7/\textbf{2.57} & \textcolor{gaingreen}{6.0} & \textcolor{gaingreen}{5.8} & \textcolor{gaingreen}{7.4} \\
\bottomrule
\end{tabular}}
\end{table*}


\subsection{End-to-end: does \ours{} improve speculative decoding?}\label{sec:headline}
Table~\ref{tab:main} compares all drafters at concurrency 1.
On both targets, \ours{} has the highest throughput on 19 of the 20 sets.

On Qwen3-8B, \ours{} is 2.6\% faster than xPress, the strongest baseline,
on the DeepSpec evaluation suite, and 9.6\% faster on the long inputs of the SPEED-Bench benchmark.
On the longest inputs, of 16K and 32K tokens, it is 12.0\% faster than xPress
and 34.6\% faster than DFlash.
Its margins over DSpark and DFlash on DeepSpec are 10.7\% and 23.6\%.
The gain over xPress does not come from higher acceptance: with 15 draft tokens,
xPress accepts slightly more tokens per step on DeepSpec (6.9 vs.\ 6.7).
It comes from a cheaper draft step, which we break down in \S\ref{sec:cost}.
On SPEED-Bench the order of acceptance flips: \ours{} accepts more tokens per step
than xPress on every input of 8K tokens or longer.

On Qwen3.8-27B, \ours{} is faster than both the official DFlash2 release and xPress
on every one of the 20 sets. On DeepSpec, its throughput is 7.6\% higher than that of DFlash2,
5.0\% higher than xPress and 10.6\% higher than DSpark; on SPEED-Bench, it is 7.4\%, 6.0\%
and 5.8\% higher. On this target, \ours{} has the highest average acceptance length
on both benchmarks.

The gain is not tied to the DFlash backbone. On P-EAGLE \cite{peagle2026}, a parallel drafter
with a different architecture, \ours{} is 6.2\% faster than a DSpark head trained on the same
backbone (Appendix~\ref{app:peagle}). \ours{}, therefore, improves parallel block drafters in general rather than to one architecture.

\begin{table*}[tbp]
\caption{AL, speedup and TPS gain on Qwen3-8B and Qwen3.8-27B at concurrency 32, $T{=}0$. Rows are category means of the sets in Table~\ref{tab:main}. All drafters use 7 draft tokens, which gave the highest TPS for each drafter in a sweep from 3 to 15 (Appendix~\ref{app:ksweep}). Columns as in Table~\ref{tab:main}.}
\label{tab:c32}
\vspace{5pt}
\centering\scriptsize\setlength{\tabcolsep}{3pt}
\resizebox{\textwidth}{!}{%
\begin{tabular}{@{}lrccccccccrccccccc@{}}
\toprule
 & \multicolumn{9}{c}{Qwen3-8B, speculation length 7} & \multicolumn{8}{c}{Qwen3.8-27B, speculation length 7} \\
\cmidrule(lr){2-10}\cmidrule(lr){11-18}
 & AR & \multicolumn{5}{c}{AL / speedup over AR ($\times$)} & \multicolumn{3}{c}{TPS gain (\%) vs} & AR & \multicolumn{4}{c}{AL / speedup over AR ($\times$)} & \multicolumn{3}{c}{TPS gain (\%) vs} \\
\cmidrule(lr){3-7}\cmidrule(lr){8-10}\cmidrule(lr){12-15}\cmidrule(lr){16-18}
 & TPS & DFlash & \textcolor{gray}{DFlash2$^\dagger$} & DSpark & xPress & \ours{} & xPress & DSpark & DFlash & TPS & DFlash2 & DSpark & xPress & \ours{} & xPress & DSpark & DFlash2 \\
\midrule
\multicolumn{18}{@{}l}{\textbf{DeepSpec}} \\
\hspace{3pt}Math & 3551 & 5.4/3.14 & \textcolor{gray}{5.6/3.28} & 6.2/3.54 & 5.9/3.39 & 6.1/\textbf{3.64} & \textcolor{gaingreen}{7.3} & \textcolor{gaingreen}{2.8} & \textcolor{gaingreen}{16.2} & 1696 & 4.6/2.26 & 4.5/2.19 & 4.7/2.25 & 4.9/\textbf{2.33} & \textcolor{gaingreen}{3.6} & \textcolor{gaingreen}{6.0} & \textcolor{gaingreen}{3.0} \\
\hspace{3pt}Code & 3364 & 4.9/2.91 & \textcolor{gray}{4.7/2.77} & 5.5/3.06 & 5.2/2.96 & 5.5/\textbf{3.23} & \textcolor{gaingreen}{9.6} & \textcolor{gaingreen}{5.4} & \textcolor{gaingreen}{11.1} & 1767 & 4.1/2.17 & 3.9/2.07 & 4.2/2.16 & 4.3/\textbf{2.25} & \textcolor{gaingreen}{4.3} & \textcolor{gaingreen}{8.7} & \textcolor{gaingreen}{3.9} \\
\hspace{3pt}Chat & 3287 & 2.9/1.87 & \textcolor{gray}{3.4/2.14} & 3.7/\textbf{2.30} & 3.4/2.08 & 3.6/2.26 & \textcolor{gaingreen}{9.0} & \textcolor{lossred}{-1.3} & \textcolor{gaingreen}{21.0} & 1636 & 3.8/2.02 & 3.6/1.91 & 3.8/1.98 & 3.9/\textbf{2.04} & \textcolor{gaingreen}{2.7} & \textcolor{gaingreen}{6.8} & \textcolor{gaingreen}{0.8} \\
\hspace{3pt}Agent & 3387 & 3.2/2.07 & \textcolor{gray}{3.5/2.23} & 3.9/2.39 & 3.6/2.22 & 3.9/\textbf{2.44} & \textcolor{gaingreen}{9.9} & \textcolor{gaingreen}{2.2} & \textcolor{gaingreen}{18.1} & 1908 & 4.5/2.31 & 4.2/2.19 & 4.6/2.30 & 4.6/\textbf{2.39} & \textcolor{gaingreen}{3.9} & \textcolor{gaingreen}{9.1} & \textcolor{gaingreen}{3.5} \\
\cmidrule(lr){1-18}
\textit{Average} & 3409 & 4.4/2.66 & \textcolor{gray}{4.6/2.74} & 5.1/2.98 & 4.8/2.82 & 5.1/\textbf{3.06} & \textcolor{gaingreen}{8.6} & \textcolor{gaingreen}{2.9} & \textcolor{gaingreen}{15.3} & 1722 & 4.2/2.17 & 4.1/2.08 & 4.3/2.16 & 4.4/\textbf{2.24} & \textcolor{gaingreen}{3.6} & \textcolor{gaingreen}{7.4} & \textcolor{gaingreen}{2.8} \\
\midrule
\multicolumn{18}{@{}l}{\textbf{SPEED-Bench long context}} \\
\hspace{3pt}Code \& sorting & 858 & 2.5/1.42 & \textcolor{gray}{3.5/1.61} & 3.4/1.62 & 2.9/1.53 & 3.5/\textbf{1.68} & \textcolor{gaingreen}{7.5} & \textcolor{gaingreen}{3.8} & \textcolor{gaingreen}{12.1} & 948 & 3.9/1.34 & 4.2/1.32 & 4.0/1.33 & 4.1/\textbf{1.37} & \textcolor{gaingreen}{3.2} & \textcolor{gaingreen}{3.8} & \textcolor{gaingreen}{2.2} \\
\hspace{3pt}Books \& writing & 1084 & 1.9/1.22 & \textcolor{gray}{2.8/1.52} & 2.5/1.41 & 2.2/1.29 & 2.6/\textbf{1.44} & \textcolor{gaingreen}{4.9} & \textcolor{lossred}{-2.1} & \textcolor{gaingreen}{8.9} & 985 & 3.0/\textbf{1.17} & 2.9/1.13 & 3.0/1.15 & 3.1/1.16 & \textcolor{gaingreen}{0.9} & \textcolor{gaingreen}{2.2} & \textcolor{lossred}{-0.9} \\
\cmidrule(lr){1-18}
\textit{Average} & 971 & 2.2/1.32 & \textcolor{gray}{3.1/1.57} & 2.9/1.52 & 2.5/1.41 & 3.0/\textbf{1.56} & \textcolor{gaingreen}{6.2} & \textcolor{gaingreen}{0.8} & \textcolor{gaingreen}{10.5} & 966 & 3.5/1.25 & 3.5/1.22 & 3.5/1.24 & 3.6/\textbf{1.26} & \textcolor{gaingreen}{2.1} & \textcolor{gaingreen}{3.1} & \textcolor{gaingreen}{0.8} \\
\bottomrule
\end{tabular}}
\end{table*}
\subsection{Serving scale: what happens as concurrency grows?}\label{sec:conc}
At high load, a shorter speculation length is faster: on Qwen3-8B at concurrency 32,
7 draft tokens gives the highest throughput for every drafter in a sweep from 3 to 15
(Appendix~\ref{app:ksweep}). Table~\ref{tab:c32} therefore compares all drafters with
7 draft tokens at concurrency 32. \ours{} remains the fastest of the compared drafters
on average, on both targets and both benchmarks. On Qwen3-8B, its throughput on DeepSpec
is 8.6\%, 2.9\% and 15.3\% higher than that of xPress, DSpark and DFlash; on Qwen3.8-27B,
it is 3.6\%, 7.4\% and 2.8\% higher than that of xPress, DSpark and DFlash2.

These margins follow from how \ours{} saves time. It can be faster in two ways: its draft
step is cheaper, and it can accept more tokens per step. At high load, verification takes
most of each step, so a cheaper draft step saves only a small share of the time, whereas
each extra accepted token still saves part of a verification step. Against DSpark, \ours{}
has only the first advantage: with 7 draft tokens, DSpark's sequential head is short and its
acceptance matches ours (5.1 tokens per step on DeepSpec), so the margin is small. Against
xPress, it has both: with 7 draft tokens, \ours{} accepts more tokens per step (5.1 vs.\ 4.8),
so the margin is larger than at concurrency 1. On Qwen3.8-27B, \ours{} accepts more tokens
per step than every baseline, which keeps it ahead of all of them.

\begin{figure}[tbp]
\centering\IfFileExists{figures/fig_conc_k7_xin_1030.pdf}{\includegraphics[width=0.9\columnwidth]{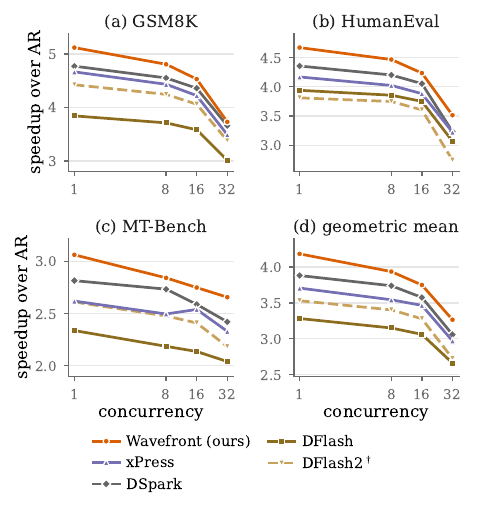}}{\fbox{\small upload figures/fig\_conc\_k7.pdf}}
\caption{Speedup over autoregressive decoding vs.\ concurrency (Qwen3-8B, 7 draft tokens, mean of two runs): GSM8K, HumanEval,
MT-Bench and their geometric mean; numbers in Table~\ref{tab:conc}. $^\dagger$As in Table~\ref{tab:main}.}
\label{fig:conc}
\end{figure}

Figure~\ref{fig:conc} follows GSM8K, HumanEval and MT-Bench from concurrency 1 to 32.
\ours{} has the highest throughput in all twelve set--concurrency cells, with an average gain
of 8.1--12.2\% over xPress and 4.6--7.6\% over DSpark across the four loads. As load grows,
every drafter's speedup over autoregressive decoding falls, from 5.12$\times$ to 3.73$\times$
for \ours{} on GSM8K, because verification takes a larger share of each step.

\subsection{Ablation: how small can the base be?}\label{sec:abl}

\begin{table}[!tb]

\caption{AL / TPS of the same causal head on bases of different rank (Qwen3-8B, DFlash backbone, 7 draft tokens, concurrency 1, $T{=}0$, $P{=}5$, mean of two runs), best TPS per row in bold; gain: TPS gain of the rank-512 base.}
\label{tab:rank}
\centering\scriptsize\setlength{\tabcolsep}{1.2pt}
\resizebox{\columnwidth}{!}{\begin{tabular}{@{}lcccccc@{}}
\toprule
 & \multicolumn{4}{c}{AL / TPS} & \multicolumn{2}{c}{gain (\%) vs} \\
\cmidrule(lr){2-5}\cmidrule(lr){6-7}
 & DSpark & r256 & r512 & full-rank & DSpark & full-rank \\
\midrule
\hspace{3pt}GSM8K & 6.36/710 & 6.38/\textbf{765} & 6.43/764 & 6.45/731 & \textcolor{gaingreen}{7.6} & \textcolor{gaingreen}{4.5} \\
\hspace{3pt}MATH-500 & 6.29/719 & 6.31/\textbf{775} & 6.35/773 & 6.35/737 & \textcolor{gaingreen}{7.5} & \textcolor{gaingreen}{4.9} \\
\hspace{3pt}AIME 2025 & 5.71/657 & 5.68/703 & 5.79/\textbf{710} & 5.74/671 & \textcolor{gaingreen}{8.1} & \textcolor{gaingreen}{5.8} \\
\hspace{3pt}HumanEval & 5.67/643 & 5.65/687 & 5.78/\textbf{696} & 5.77/663 & \textcolor{gaingreen}{8.2} & \textcolor{gaingreen}{5.0} \\
\hspace{3pt}MBPP & 5.30/599 & 5.33/\textbf{646} & 5.33/640 & 5.40/620 & \textcolor{gaingreen}{6.8} & \textcolor{gaingreen}{3.2} \\
\hspace{3pt}LiveCodeBench & 5.71/625 & 5.69/667 & 5.77/\textbf{671} & 5.74/638 & \textcolor{gaingreen}{7.4} & \textcolor{gaingreen}{5.2} \\
\hspace{3pt}MT-Bench & 3.71/425 & 3.62/447 & 3.70/\textbf{452} & 3.76/438 & \textcolor{gaingreen}{6.4} & \textcolor{gaingreen}{3.2} \\
\hspace{3pt}Alpaca & 3.61/413 & 3.49/429 & 3.64/\textbf{442} & 3.63/421 & \textcolor{gaingreen}{7.0} & \textcolor{gaingreen}{5.0} \\
\cmidrule(lr){1-7}
\textit{Average} & 5.29/599 & 5.27/640 & 5.35/\textbf{644} & 5.36/615 & \textcolor{gaingreen}{7.5} & \textcolor{gaingreen}{4.7} \\
\bottomrule
\end{tabular}}
\end{table}

Under the same chain-aware head on Qwen3-8B (7 draft tokens, concurrency 1), a rank-512 base matches the acceptance length of the
full lm\_head (5.35 vs.\ 5.36 on average) and is 4.7\% faster, because its readout is cheaper; rank 256 is no faster and accepts
0.08 fewer tokens (Table~\ref{tab:rank}). On the P-EAGLE backbone, the full-rank base likewise gains
0.09 AL but is 2.9\% slower than rank 512 (Appendix~\ref{app:peagle}).

\begin{figure}[tbp]
\centering\IfFileExists{figures/fig_passes_k7.pdf}{\includegraphics[width=\columnwidth]{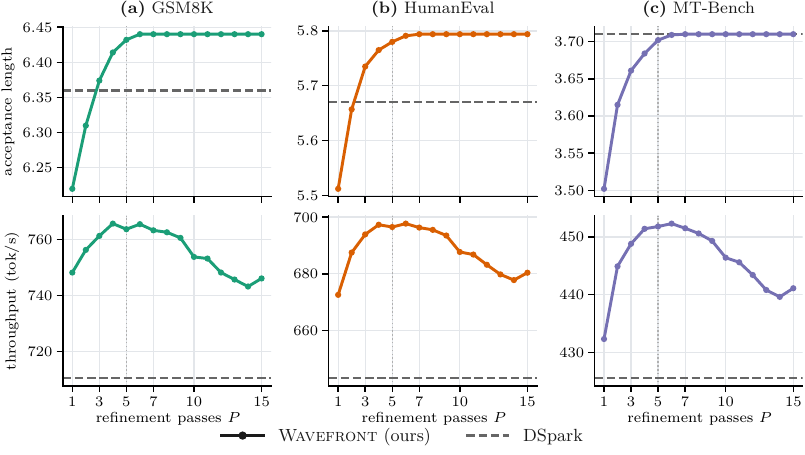}}{\fbox{\small upload figures/fig\_passes\_k7.pdf}}
\caption{\ours{} vs.\ the number of refinement passes $P$ on Qwen3-8B (rank-512 base, concurrency 1, $T{=}0$, speculation length 7), one column per
set: top, acceptance length; bottom, throughput.}
\label{fig:passes}
\end{figure}

\subsection{Acceptance and throughput vs.\ refinement passes}\label{sec:passes}
Sweeping the number of refinement passes on Qwen3-8B (7 draft tokens, concurrency 1; Fig.~\ref{fig:passes}), AL stops improving once $P$ reaches the block length ($P{=}6$--7) and throughput peaks at
$P{=}4$--6, so we use $P{=}5$; even $P{=}1$ is faster than DSpark on GSM8K, HumanEval and MT-Bench.

\subsection{Where does the time go?}\label{sec:cost}
\begin{figure}[!tb]
\centering\IfFileExists{figures/fig_cost_k7.pdf}{\includegraphics[width=0.98\columnwidth]{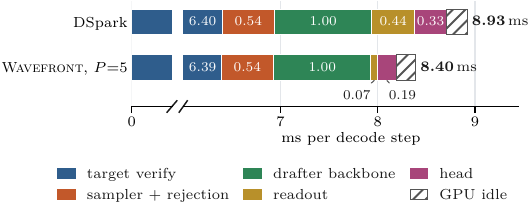}}{\fbox{\small upload figures/fig\_cost\_k7.pdf}}
\caption{The stages of one decode step in execution order (Qwen3-8B, DFlash backbone, ms per decode step, concurrency 1, GSM8K, speculation length 7).}
\label{fig:cost}
\end{figure}

Figure~\ref{fig:cost} splits one decode step on Qwen3-8B (GSM8K, concurrency 1, 7 draft tokens) into stages
(Table~\ref{tab:cost} in Appendix~\ref{app:extra}). Verification, sampling and the drafter backbone cost the same for both
drafters; \ours{} saves in the readout (0.07 instead of 0.44\,ms) and the head (five parallel passes, 0.19\,ms, against 0.33\,ms
for DSpark's sequential head), so a step takes 8.40 instead of 8.93\,ms.

\section{Related work}\label{sec:related}
\paragraph{Speculative decoding.} Speculative decoding lets a cheaper model draft several tokens that the
target verifies in a single forward pass; rejection sampling keeps the target's output distribution
unchanged \cite{leviathan2023,chen2023}. Learned drafters improve the draft by reading the target's own
representations. Medusa adds extra decoding heads that predict several future tokens in parallel, each
independently of the others \cite{medusa}, and Hydra makes these heads sequentially dependent by feeding each
head the tokens drafted before it \cite{ankner2024hydra}. EAGLE drafts autoregressively at the feature level
\cite{eagle}, EAGLE-2 organizes the draft into dynamic trees \cite{eagle2}, and EAGLE-3 predicts tokens
directly from fused multi-layer features \cite{eagle3}. These autoregressive drafters still spend one drafter
pass per token.

\paragraph{Parallel drafters.} Parallel drafters remove this loop by drafting a whole block at once.
ParallelSpec appends mask tokens to the input and predicts several future tokens from them in one pass
\cite{xiao2024parallelspec}, and PARD adapts an autoregressive draft model to predict multiple future tokens
in a single forward pass, with one draft model serving a whole family of targets \cite{an2026pard}. DFlash
uses a lightweight block-diffusion model that fills a block of masked slots in a single forward pass,
conditioned on features from several target layers that are injected into the keys and values of every draft
layer \cite{dflash2026}. P-EAGLE drafts all positions of an EAGLE-style drafter in one pass, replacing the
unknown tokens and hidden states with a learned mask embedding and a shared hidden state, and trains on long
sequences with PARD's conditional token dropping \cite{peagle2026}. Drafting in parallel, however, removes the
dependence of each slot on the tokens drafted before it.

\paragraph{Restoring dependence inside the block.} Recent methods restore this dependence in three ways. The
first corrects the block sequentially. DSpark adds a rank-256 previous-token Markov bias to the backbone's base
logits and samples the block from left to right \cite{dspark2026}. Domino runs a lightweight GRU over the
preceding draft tokens and adds a low-rank logit correction at each slot, again from left to right
\cite{huang2026domino}. The second keeps several candidates per slot and selects among them: DFlash2 adds
two-tap dynamic convolutions to the drafter trunk and selects a path through the top-16 candidates at each slot
with a low-rank pairwise scorer \cite{inco2026dflash2}. The third refines the block in parallel. xPress
\cite{wang2026xpress}, concurrent with our work, adds a causal refiner that mixes slots through a learned
lower-triangular matrix per channel, whose weights are fixed after training, and resolves the block over about
six parallel Jacobi passes. DSpark, Domino and xPress all add their correction to base logits read from the
target's full lm\_head. \ours{} also refines in parallel, but it replaces the full head with a learned low-rank
base and mixes slots with attention whose weights depend on the drafted tokens.

\paragraph{Cheaper drafter readouts.} With large vocabularies, the drafter's readout to the vocabulary becomes
a significant share of draft time. FR-Spec restricts the drafter's lm\_head to a frequency-ranked subset of the
vocabulary, reducing its computation by 75\% when keeping 32K of Llama-3-8B's 128K tokens \cite{zhao2025frspec}.
\ours{} reduces the same cost differently: it keeps the full vocabulary and learns a low-rank base instead, and
\S\ref{sec:seed} shows that this base gives the correction head the same seed as the full lm\_head.

\paragraph{Parallel decoding.} Jacobi decoding treats a causal sequence as a system of equations and solves it
by updating all positions in parallel; it returns exactly the sequential result in at most as many iterations
as there are positions \cite{song2021,santilli2023}. Lookahead decoding applies Jacobi iteration to the target
model itself and verifies candidate $n$-grams collected along the way \cite{fu2024}, and CLLMs fine-tune the
target so that Jacobi decoding converges in fewer iterations \cite{kou2024cllms}. \ours{} uses Jacobi iteration
more narrowly: it iterates a small drafter-side head over one draft block, for a fixed number of passes.

\paragraph{Serving systems.} Other work changes how verification is scheduled rather than how the draft is produced; DSpark,
for example, adapts the verification length of each request to its estimated acceptance \cite{dspark2026}, a change complementary
to \ours{}. Our implementation follows the constraints of production serving in vLLM \cite{vllm}: fixed shape buckets, CUDA-graph
capture, and no large vocabulary-sized tensors.

\section{Conclusion}
Parallel block drafting removes sequential work from the drafter, yet existing correction heads
still read the full lm\_head for a base prediction that the head often rewrites. \ours{} therefore pairs a cheap low-rank base, which provides the seed, with a chain-aware head that corrects it over a few
parallel passes. On
Qwen3-8B and Qwen3.8-27B, \ours{} has the highest average throughput of the compared parallel
drafters at both low and high load, and its gain carries over to different backbones. More
broadly, our results suggest that drafter design should ask not only how much capacity to use, but
\emph{where to put it}.

\bibliographystyle{mlsys2025}
\bibliography{paper}

\section*{AI Usage Disclosure}
The authors used generative AI tools as follows.
\begin{itemize}
\item \textbf{Tools:} Claude (Anthropic) and ChatGPT (OpenAI).
\item \textbf{Ideation and experiment design:} The research questions, system design and
experimental plan were developed by the authors without AI assistance.
\item \textbf{Literature review:} AI assistants helped locate related work, check citations
against their sources and draft parts of the related-work discussion; the authors reviewed
all cited sources and verified every claim about them.
\item \textbf{Code and experiments:} AI coding assistants supported code development and
plotting scripts, which the authors reviewed and tested. No AI tool generated or modified
any experimental data or reported result.
\item \textbf{Writing and figures:} AI assisted with drafting and editing text and with
figure styling. All text and figures were revised and verified by the authors.
\end{itemize}
The authors take full responsibility for the technical content and conclusions.

\clearpage

\appendix
\section*{Appendix}

\section{Notation}\label{app:notation}
Table~\ref{tab:notation} summarizes the symbols used in the paper.

\begin{table}[!h]
\caption{Notation. Superscript $(t)$ denotes refinement pass $t$; $x_k^{(0)}$ is the seed.}
\label{tab:notation}
\vspace{5pt}
\centering\footnotesize\setlength{\tabcolsep}{4pt}
\begin{tabular}{@{}cl@{}}
\toprule
symbol & meaning \\
\midrule
$\K$ & draft slots per block \\
$V$ & vocabulary size \\
$d$ & backbone width \\
$b$ & base rank \\
$r$ & token-code width \\
$P$ & refinement passes \\
$x_0$ & verified anchor \\
$m_k$ & mask token at slot $k$ \\
$x_k^{(t)}$ & token at slot $k$ after pass $t$ \\
\midrule
$h_k$ & backbone hidden state \\
$q_k$ & low-rank feature \\
$z_k$ & normalized feature, $\mathrm{RMSNorm}(q_k)$ \\
$U_k$ & base logits \\
$c_k^{(t)}$ & predecessor-token code \\
$u_k^{(t)}$ & head latent \\
$\ell_k^{(t)}$ & corrected logits \\
\midrule
$W_z$ & projection, $b{\times}d$ \\
$W_{\mathrm{base}}$ & base readout, $V{\times}b$ \\
$W_1$ & token codes, $V{\times}r$ \\
$W_{\mathrm{in}}$ & fusion, $r{\times}(r{+}b)$ \\
$W_2$ & head readout, $V{\times}r$ \\
\bottomrule
\end{tabular}
\end{table}

\section{Implementation details}\label{app:impl}
We implement \ours{} as a native speculator in vLLM v0.32
nightly build. The backbone forward pass, the shared $W_z$ projection, the low-rank
readout and all $P$ refinement passes are captured in a single CUDA graph per batch bucket. Because $P$ is fixed, the draft step has
no data-dependent control flow, and $P$ is read once when the speculator is initialized.

\paragraph{Fused kernels.} Two fused kernels keep the head-side critical path short. The first combines the query and key
projections of the head's attention into one GEMM and applies RoPE from pre-interleaved tables. The second is a two-launch Triton
epilogue that adds the correction to the base logits and takes the argmax for each slot, so the full corrected $[BK, V]$ logit
tensor is never materialized.

\paragraph{Exactness.} Both kernels are bit-identical to the eager path. We verified this on 60 slots over 5 trials, on GPU and on
CPU against the training module. An A/B comparison of the fused and unfused serving paths, with prefix caching disabled, produced
bit-identical token, step and acceptance counters at concurrency 1 and 8.

\section{Measurement details}\label{app:serving}
\paragraph{Hardware.} Within a comparison, all drafters of a target run on the same node, one server per GPU. On Qwen3.8-27B the
drafters are split over two nodes; both nodes also run DFlash2, through which speedups are normalized.

\paragraph{Prompts.} On Qwen3-8B, the DeepSpec sets run in full, except Alpaca (first 1K prompts) and LiveCodeBench (500 prompts);
SPEED-Bench cells use 32 prompts at concurrency 1 and 128 at concurrency 32. On Qwen3.8-27B, concurrency-1 cells use the first 32
prompts of each DeepSpec set with 4096 output tokens and 16 SPEED-Bench prompts; concurrency-32 cells use the full DeepSpec sets and
64 SPEED-Bench prompts, with four runs per cell. SPEED-Bench outputs are 1024 tokens, and prompts that do not fit the 40,960-token
context with this output budget are dropped for every drafter.

\paragraph{Tensor parallelism.} On Qwen3.8-27B, every drafter takes its draft tokens from tensor-parallel rank 0. Otherwise the two
ranks occasionally draft different tokens from near-tied logits and the output degenerates, an issue of vLLM's DFlash path that
affects all compared drafters.

\paragraph{Builds and counters.} All experiments run in vLLM v0.32 nightly build that merged xPress. AL comes from vLLM's speculative-decoding
counters, and the autoregressive reference uses the same build and flags with speculation disabled.

\section{Concurrency sweep}\label{app:conc}
Table~\ref{tab:conc} gives the numbers behind Figure~\ref{fig:conc}: AL, speedup over autoregressive decoding and \ours{}'s
throughput gain on GSM8K, HumanEval and MT-Bench at concurrency 1, 8, 16 and 32 (Qwen3-8B, 7 draft tokens, mean of two runs).
\begin{table*}[!t]
\caption{AL, speedup and TPS gain on Qwen3-8B as concurrency grows, $T{=}0$, 7 draft tokens, mean of two runs. Columns as in Table~\ref{tab:main}.}
\label{tab:conc}
\centering\scriptsize\setlength{\tabcolsep}{4pt}
\begin{tabular}{@{}lrcccccccc@{}}
\toprule
 & AR & \multicolumn{5}{c}{AL / speedup over AR ($\times$)} & \multicolumn{3}{c}{TPS gain (\%) vs} \\
\cmidrule(lr){3-7}\cmidrule(lr){8-10}
 & TPS & DFlash & \textcolor{gray}{DFlash2$^\dagger$} & DSpark & xPress & \ours{} & xPress & DSpark & DFlash \\
\midrule
\multicolumn{10}{@{}l}{\textbf{Concurrency 1}} \\
\hspace{3pt}GSM8K & 149 & 4.9/3.84 & \textcolor{gray}{5.9/4.42} & 6.4/4.77 & 6.2/4.66 & 6.4/\textbf{5.12} & \textcolor{gaingreen}{9.7} & \textcolor{gaingreen}{7.3} & \textcolor{gaingreen}{33.2} \\
\hspace{3pt}HumanEval & 149 & 5.0/3.94 & \textcolor{gray}{5.0/3.81} & 5.8/4.35 & 5.4/4.17 & 5.8/\textbf{4.67} & \textcolor{gaingreen}{12.0} & \textcolor{gaingreen}{7.2} & \textcolor{gaingreen}{18.5} \\
\hspace{3pt}MT-Bench & 149 & 2.9/2.34 & \textcolor{gray}{3.4/2.61} & 3.7/2.82 & 3.4/2.62 & 3.8/\textbf{3.06} & \textcolor{gaingreen}{16.9} & \textcolor{gaingreen}{8.8} & \textcolor{gaingreen}{31.0} \\
\cmidrule(lr){1-10}
\textit{Average} & 149 & 4.3/3.37 & \textcolor{gray}{4.7/3.62} & 5.3/3.98 & 5.0/3.82 & 5.3/\textbf{4.28} & \textcolor{gaingreen}{12.2} & \textcolor{gaingreen}{7.6} & \textcolor{gaingreen}{27.0} \\
\midrule
\multicolumn{10}{@{}l}{\textbf{Concurrency 8}} \\
\hspace{3pt}GSM8K & 1155 & 4.9/3.71 & \textcolor{gray}{5.9/4.24} & 6.4/4.55 & 6.2/4.43 & 6.4/\textbf{4.81} & \textcolor{gaingreen}{8.4} & \textcolor{gaingreen}{5.6} & \textcolor{gaingreen}{29.5} \\
\hspace{3pt}HumanEval & 1102 & 5.0/3.86 & \textcolor{gray}{5.0/3.75} & 5.8/4.20 & 5.4/4.02 & 5.8/\textbf{4.47} & \textcolor{gaingreen}{11.0} & \textcolor{gaingreen}{6.3} & \textcolor{gaingreen}{15.8} \\
\hspace{3pt}MT-Bench & 1015 & 2.9/2.19 & \textcolor{gray}{3.4/2.48} & 3.8/2.73 & 3.4/2.50 & 3.7/\textbf{2.84} & \textcolor{gaingreen}{13.9} & \textcolor{gaingreen}{4.0} & \textcolor{gaingreen}{30.0} \\
\cmidrule(lr){1-10}
\textit{Average} & 1091 & 4.3/3.29 & \textcolor{gray}{4.8/3.53} & 5.3/3.87 & 5.0/3.69 & 5.3/\textbf{4.08} & \textcolor{gaingreen}{10.5} & \textcolor{gaingreen}{5.5} & \textcolor{gaingreen}{24.2} \\
\midrule
\multicolumn{10}{@{}l}{\textbf{Concurrency 16}} \\
\hspace{3pt}GSM8K & 2200 & 4.9/3.58 & \textcolor{gray}{5.9/4.06} & 6.4/4.36 & 6.2/4.22 & 6.4/\textbf{4.53} & \textcolor{gaingreen}{7.3} & \textcolor{gaingreen}{3.9} & \textcolor{gaingreen}{26.5} \\
\hspace{3pt}HumanEval & 1987 & 5.0/3.75 & \textcolor{gray}{5.0/3.61} & 5.8/4.05 & 5.4/3.88 & 5.8/\textbf{4.23} & \textcolor{gaingreen}{9.1} & \textcolor{gaingreen}{4.5} & \textcolor{gaingreen}{12.9} \\
\hspace{3pt}MT-Bench & 1685 & 2.9/2.14 & \textcolor{gray}{3.4/2.41} & 3.8/2.59 & 3.4/2.54 & 3.7/\textbf{2.75} & \textcolor{gaingreen}{8.2} & \textcolor{gaingreen}{6.2} & \textcolor{gaingreen}{28.7} \\
\cmidrule(lr){1-10}
\textit{Average} & 1957 & 4.3/3.22 & \textcolor{gray}{4.8/3.43} & 5.3/3.75 & 5.0/3.62 & 5.3/\textbf{3.92} & \textcolor{gaingreen}{8.1} & \textcolor{gaingreen}{4.6} & \textcolor{gaingreen}{21.5} \\
\midrule
\multicolumn{10}{@{}l}{\textbf{Concurrency 32}} \\
\hspace{3pt}GSM8K & 4010 & 4.9/3.01 & \textcolor{gray}{5.9/3.38} & 6.4/3.66 & 6.2/3.49 & 6.4/\textbf{3.73} & \textcolor{gaingreen}{6.9} & \textcolor{gaingreen}{1.9} & \textcolor{gaingreen}{23.8} \\
\hspace{3pt}HumanEval & 3551 & 5.0/3.07 & \textcolor{gray}{5.0/2.74} & 5.8/3.23 & 5.4/3.22 & 5.8/\textbf{3.51} & \textcolor{gaingreen}{9.0} & \textcolor{gaingreen}{8.7} & \textcolor{gaingreen}{14.5} \\
\hspace{3pt}MT-Bench & 2343 & 3.0/2.04 & \textcolor{gray}{3.4/2.18} & 3.7/2.42 & 3.4/2.33 & 3.8/\textbf{2.66} & \textcolor{gaingreen}{14.0} & \textcolor{gaingreen}{9.8} & \textcolor{gaingreen}{30.2} \\
\cmidrule(lr){1-10}
\textit{Average} & 3301 & 4.3/2.80 & \textcolor{gray}{4.7/2.87} & 5.3/3.21 & 5.0/3.12 & 5.4/\textbf{3.40} & \textcolor{gaingreen}{8.9} & \textcolor{gaingreen}{5.8} & \textcolor{gaingreen}{21.3} \\
\bottomrule
\end{tabular}
\end{table*}

\section{P-EAGLE backbone}\label{app:peagle}

\begin{table}[!tb]
\caption{AL / TPS on Qwen3-8B with the P-EAGLE backbone (7 draft tokens, concurrency 1, $T{=}0$,
$P{=}5$); best TPS per row in bold. Gain: TPS gain over the DSpark head. Full-rank: \ours{} with the full lm\_head as its base.}
\label{tab:peagle}
\centering\scriptsize\setlength{\tabcolsep}{2.2pt}
\resizebox{\columnwidth}{!}{\begin{tabular}{@{}lccccc@{}}
\toprule
 & \multicolumn{3}{c}{AL / TPS} & \multicolumn{2}{c}{gain (\%)} \\
\cmidrule(lr){2-4}\cmidrule(lr){5-6}
 & DSpark & \ours{} & full-rank & \ours{} & full-rank \\
\midrule
\hspace{3pt}GSM8K & 6.3/685 & 6.4/\textbf{730} & 6.5/704 & \textcolor{gaingreen}{6.6} & \textcolor{gaingreen}{2.8} \\
\hspace{3pt}MATH-500 & 6.2/693 & 6.3/\textbf{740} & 6.3/710 & \textcolor{gaingreen}{6.8} & \textcolor{gaingreen}{2.5} \\
\hspace{3pt}AIME 2025 & 5.8/642 & 5.8/\textbf{685} & 5.8/661 & \textcolor{gaingreen}{6.7} & \textcolor{gaingreen}{3.0} \\
\hspace{3pt}HumanEval & 5.7/621 & 5.7/\textbf{658} & 5.8/643 & \textcolor{gaingreen}{6.0} & \textcolor{gaingreen}{3.5} \\
\hspace{3pt}MBPP & 5.3/576 & 5.3/\textbf{614} & 5.4/594 & \textcolor{gaingreen}{6.6} & \textcolor{gaingreen}{3.1} \\
\hspace{3pt}LiveCodeBench & 5.7/605 & 5.7/\textbf{634} & 5.8/619 & \textcolor{gaingreen}{4.8} & \textcolor{gaingreen}{2.3} \\
\hspace{3pt}MT-Bench & 3.9/434 & 3.9/\textbf{458} & 4.0/447 & \textcolor{gaingreen}{5.5} & \textcolor{gaingreen}{3.0} \\
\hspace{3pt}Alpaca & 3.9/425 & 3.9/\textbf{450} & 4.0/446 & \textcolor{gaingreen}{5.9} & \textcolor{gaingreen}{4.9} \\
\cmidrule(lr){1-6}
\textit{Average} & 5.3/585 & 5.4/\textbf{621} & 5.5/603 & \textcolor{gaingreen}{6.2} & \textcolor{gaingreen}{3.1} \\
\bottomrule
\end{tabular}}
\end{table}

To test whether \ours{} depends on the DFlash backbone, we apply it to P-EAGLE \cite{peagle2026},
a parallel drafter with a different architecture. The P-EAGLE backbone has five draft layers,
reads the same target layers $\{1,9,17,25,33\}$ as the DFlash backbone, and is trained on the same
data with geometric conditional-drop subsampling (0.7), for 10 epochs at batch size 64. As in the
controlled ablations, the DSpark head, \ours{} and its full-rank variant share this backbone, its
data and its schedule.

Table~\ref{tab:peagle} reports the results on Qwen3-8B with 7 draft tokens at concurrency 1.
\ours{} has the highest throughput on all eight sets, 4.8--6.8\% higher than the DSpark head and
6.2\% higher on average, at the same average acceptance length. The full-rank base gains 0.09 AL
on average but runs 2.9\% slower than the rank-512 base, the same trade-off as on the DFlash
backbone (\S\ref{sec:abl}).

\section{Speculation length at concurrency 32}\label{app:ksweep}
Table~\ref{tab:ksweep} sweeps the speculation length at concurrency 32 on the setup of Table~\ref{tab:c32} ($T{=}0$, the twelve
DeepSpec sets, three runs per cell). Seven tokens is the fastest length for every drafter: by 4.8\% over nine for xPress, and within
run-to-run noise of nine for \ours{} and DSpark; fifteen is 5.7\% (xPress) to 13\% (\ours{}, DSpark) slower.
\begin{table}[!tb]
\caption{Qwen3-8B, concurrency 32, $T{=}0$: TPS (geometric mean over the twelve DeepSpec sets, mean of three runs) for each
speculation length; fastest per row in bold.}
\label{tab:ksweep}
\centering\scriptsize\setlength{\tabcolsep}{3pt}
\begin{tabular}{@{}lcccccc@{}}
\toprule
 & \multicolumn{6}{c}{speculation length} \\ \cmidrule(lr){2-7}
 & 3 & 5 & 7 & 9 & 11 & 15 \\
\midrule
xPress & 7627 & 8805 & \textbf{9293} & 8871 & 8296 & 8765 \\
\ours{} & 7861 & 9310 & \textbf{10068} & 10004 & 9674 & 8746 \\
DSpark & 7864 & 9307 & \textbf{9824} & 9771 & 9350 & 8517 \\
\bottomrule
\end{tabular}
\end{table}

\section{Additional results}\label{app:diag}\label{app:extra}
This appendix gives the detailed measurements behind two claims in the main text.
Table~\ref{tab:slotacc} supports \S\ref{sec:seed}: it reports, at selected slots, how often
the base's top token matches the target's token before any refinement pass, for the full
lm\_head and for low-rank bases of rank 512 and 256 on GSM8K, HumanEval and MT-Bench. All
three bases give the same seed quality, which is why a low-rank base suffices.
Table~\ref{tab:cost} supports \S\ref{sec:cost}: it breaks one decode step on Qwen3-8B into
GPU kernel time per stage, and gives the numbers behind Figure~\ref{fig:cost}. It shows that
\ours{} saves time only in the readout and the head, while verification, sampling and the
drafter backbone cost the same as for DSpark.



\begin{table}[!tb]
\caption{Top-1 accuracy of the base alone, before any refinement pass, at selected slots $k$ ($\K{=}15$, teacher-forced; same
backbone and head, three bases).}
\label{tab:slotacc}
\vspace{5pt}
\centering\footnotesize\setlength{\tabcolsep}{3.5pt}
\begin{tabular}{llccccc}
\toprule
dataset & base & $k{=}1$ & 5 & 10 & 15 & mean \\
\midrule
\multirow{3}{*}{GSM8K}     & full rank & 0.963 & 0.810 & 0.643 & 0.483 & 0.714 \\
                           & rank 512  & 0.962 & 0.809 & 0.638 & 0.475 & 0.710 \\
                           & rank 256  & 0.962 & 0.811 & 0.644 & 0.488 & 0.715 \\
\midrule
\multirow{3}{*}{HumanEval} & full rank & 0.947 & 0.739 & 0.530 & 0.354 & 0.625 \\
                           & rank 512  & 0.945 & 0.736 & 0.526 & 0.343 & 0.620 \\
                           & rank 256  & 0.945 & 0.736 & 0.532 & 0.355 & 0.625 \\
\midrule
\multirow{3}{*}{MT-Bench}  & full rank & 0.845 & 0.471 & 0.279 & 0.177 & 0.398 \\
                           & rank 512  & 0.841 & 0.474 & 0.278 & 0.173 & 0.396 \\
                           & rank 256  & 0.834 & 0.469 & 0.279 & 0.174 & 0.395 \\
\bottomrule
\end{tabular}
\end{table}

\begin{table}[!tb]
\caption{GPU kernel time per decode step by stage (ms; Qwen3-8B, DFlash backbone, concurrency 1, GSM8K-500, $T{=}0$, speculation length 7). The step is the wall time per target verification; idle is the step minus the kernel time.}
\label{tab:cost}
\centering\scriptsize\setlength{\tabcolsep}{3pt}
\begin{tabular}{lcc}
\toprule
stage (ms/step) & DSpark & \ours{} $P{=}5$ \\
\midrule
target verify (+lm\_head) & 6.40 & 6.39 \\
target sampler+rejection & 0.54 & 0.54 \\
drafter backbone (+ctx KV) & 1.00 & 1.00 \\
readout GEMM & 0.44 & \textbf{0.07} \\
head & 0.33 & \textbf{0.19} \\
\midrule
GPU busy & 8.72 & \textbf{8.21} \\
GPU idle & 0.21 & 0.20 \\
step & 8.93 & \textbf{8.40} \\
\bottomrule
\end{tabular}
\end{table}

\end{document}